\documentclass[letterpaper]{article} 
\usepackage[]{aaai25}  
\usepackage{times}  
\usepackage{helvet}  
\usepackage{courier}  
\usepackage[hyphens]{url}  
\usepackage{graphicx} 
\usepackage{natbib}  
\usepackage{caption} 
\usepackage{algorithm}
\usepackage{algorithmic}
\usepackage{booktabs}
\usepackage{graphicx}
\usepackage{tabularx}
\usepackage{array}
\usepackage{xcolor}
\newcolumntype{H}{>{\setbox0=\hbox\bgroup}c<{\egroup}@{}}

\usepackage{newfloat}
\usepackage{listings}
\DeclareCaptionStyle{ruled}{labelfont=normalfont,labelsep=colon,strut=off} 
\floatstyle{ruled}
\newfloat{listing}{tb}{lst}{}
\floatname{listing}{Listing}
\title{PPLqa: An Unsupervised Information-Theoretic Quality Metric for\\ Comparing Generative Large Language Models}

\author{Gerald Friedland\textsuperscript{\rm 1}, Xin Huang\textsuperscript{\rm 1}, Yueying Cui\textsuperscript{\rm 1},   
Vishaal Kapoor\textsuperscript{\rm 1},  Ashish Khetan\textsuperscript{\rm 1}, Sanjiv Das\textsuperscript{\rm 1,2} \\ 
}

\affiliations{
    \textsuperscript{\rm 1}Amazon Web Services\thanks{We are grateful to Patrick Haffner and Pola Schw{\"o}bel for many useful suggestions. },
    \textsuperscript{\rm 2}Santa Clara University

}

\usepackage{bibentry}

\begin{document}

\maketitle

\begin{abstract}
We propose PPLqa, an easy to compute, language independent, information-theoretic metric to measure the quality of responses of generative Large Language Models (LLMs) in an unsupervised way, without requiring  ground truth annotations or human supervision. The method and metric enables users to rank generative language models for quality of responses, so as to make a selection of the best model for a given task. Our single metric assesses LLMs with an approach that subsumes, but is not explicitly based on, coherence and fluency (quality of writing) and relevance and consistency (appropriateness of response) to the query. PPLqa performs as well as other related metrics, and works better with long-form Q\&A. Thus, PPLqa enables bypassing the lengthy annotation process required for ground truth evaluations, and it also correlates well with  human and LLM rankings. 
\end{abstract}

%

\section{Introduction}
\label{sec:intro}
The strength of generative Large Language Models (LLMs) lies in answering questions (prompts),  smoothly reorganizing existing documents (as in summarization), or generating new ones, emulating a human narrator at various levels of intellectual depth. Of course, the level of intellectual depth is limited by the match between training and task. The problem that this article examines is the comparison of the quality of various responses, for a given task, across various Large Language Models.

Current work\footnote{\url{https://hai.stanford.edu/news/language-models-are-changing-ai-we-need-understand-them}}, such as HELM (\citet{liang_holistic_2022}), Eleuther AI LM Harness \citep{gao_framework_2021}, Big-BENCH \citep{srivastava_beyond_2022}, focuses mainly on computing metrics against ground truth.\footnote{\url{https://github.com/aws-samples/llm-evaluation-methodology}} 
Not only is domain-specific ground truth preparation a costly process, it is also inherently difficult as it is the equivalent of creating exam question and answers -- a task that is usually best performed by experts with relevant domain understanding.

Assuming the presence of ground truth, one still requires an objective loss function to compare the output of the model bit-by-bit with the answers prepared by the experts. Furthermore, practice has shown \citep{lin_llm-eval_2023} that given the broad applicability of these la nguage models, it can be non-trivial to specify a sufficient test and validation set.   Enterprise use of generative language models therefore either requires blind trust in the LLM or the acceptance of the risk that the language model may have surprising responses to some prompts. The GPT-4 technical report \cite{openai_gpt-4_2023} qualifies these challenges as follows: ``Despite its capabilities, GPT-4 has similar limitations to earlier GPT models: it is not fully reliable (e.g. can suffer from “hallucinations”), has a limited context window, and does not learn from experience. Care should be taken when using the outputs of GPT-4, particularly in contexts where reliability is important." Despite large increases in context size, the hallucination problem remains. 

Therefore, in this paper, we explore an alternative approach to evaluating LLMs using an unsupervised approach that does not require ground truth. Our method is based on an information-theoretic quantification of the structure of a prompt/response pair. Such a quantification can obviously only be very coarse. However, our evaluations show that our metrics perform in the same range as human and LLM-based evaluations. Details are presented in Section \ref{sec:theory}.

The main elements and findings of this paper are as follows.
After a short summary of related work in Section~\ref{sec:rel_work}, Section~\ref{sec:theory} explains the theoretical underpinnings of the PPLqa metric. Section~\ref{sec:experiments} presents the various analyses undertaken using the PPLqa metric. Responses are generated from a few LLMs 
to various questions in four subject domains: macroeconomics, astronomy, AI, electronics. Responses are ranked using the metric. The correlation of these rankings to human rankings is assessed using a modified version of Kendall's rank correlation $\tau$. Empirical assessments (F1, accuracy, MCC) for all four domains evidence that PPLqa and human rankings are aligned. For robustness, ranking of LLM responses are also computed on the {\tt MT-Bench} dataset, showing that PPLqa outperforms the comparison metrics. In yet another robustness test, we replaced human rankings by those from Anthropic's Claude 3 LLM -- here alignment of PPLqa with the LLM's rankings are stronger than with human rankings. There is ranking alignment across PPLqa rankings, Claude's rankings, and human rankings. The limitations of the approach are discussed in Section~\ref{sec:limits}. Concluding comments are offered in Section \ref{sec:concl}. 

\section{Related Work}
\label{sec:rel_work}

We may define an evaluation as a task used to evaluate the quality of a system's behavior. Behavior is necessarily task-related and may be measured around various categories such as over-refusals, safety, system message steerability, hallucinations, math/logical reasoning, question-answering, completions, etc. Indeed, simple choice responses are the subject of model-graded evaluations using templates. Evaluation templates consider various metrics such as (i) factual consistency with a reference answer, (ii) relevance to the prompt query, (iii) conciseness, i.e., does not contain irrelevant information, (iv) correctness, i.e., arrives at the right response. Of course, all of these metrics require ground truth, see for example \citet{wang_evaluating_2024}, or human in the loop evaluations \citet{yadav_evalai_2019}. 

We mention two families of metrics/benchmarks that can be used to assess the quality of an LLM but that differ from our PPLqa in that they require ground truth. The first family of metrics, BLEU \citep{papineni_bleu_2002}, ROUGE \citep{lin_rouge_2004}, METEOR \citep{banerjee_meteor_2005}, CIDEr \citep{vedantam_cider_2015}, and BERTScore \citep{zhang_bertscore_2020} have been used to evaluate quality in machine learning and text summarization tasks. These methods fundamentally aim to quantify the similarity between generated text and a reference text using means such as considering overlap of n-grams, word pairs, or synonyms, and weighting importance based on TF-IDF vectors. BERTScore is a similar metric that computes similarity using contextual embeddings and cosine similarity. The second family of benchmarks, GLUE \citep{wang_glue_2018} and SuperGLUE \citep{wang_superglue_2019}, are a comprehensive suite of metrics to assess the quality of LLMs on a variety of NLP tasks and benchmark datasets. These metrics are calculated automatically based on ground truth which includes correct true/false and multiple-choice responses and classifications, pronoun resolutions as well as reference texts.  

The RAGAS framework (Retrieval Augmented Generation Assessment) is designed to evaluate RAG-LLM (Retrieval Augmented Generation) pipelines \citep{es-etal-2024-ragas}. It distinguishes itself from classical metrics such as grammatical accuracy and fluency and correctness metrics such as ROUGE and BLEU with component-wise metrics such as Faithfulness, Answer Relevancy, Context Recall, Context Precision, and Context Relevancy; and end-to-end evaluation metrics such as Answer Semantic Similarity and Answer Correctness. These metrics can be evaluated using human judgement or with automated evaluation using LLMs \citep{zheng_judging_2023}, however often human oversight is required for the latter case to ensure sufficient quality of LLM evaluation.\footnote{\url{https://huggingface.co/blog/zero-shot-eval-on-the-hub}} 

Evaluations of generated text for toxicity and polarity towards a demographic group are handled already by approaches that do not need information-theoretic metrics.\footnote{ \url{https://huggingface.co/blog/evaluating-llm-bias}} Privacy is another issue with LLMs revealing PII in the training data. Parallel issues around copyright on data used for training also arise.

Related work that does not require expensive ground truth is \citet{fu_gptscore_2023}, which introduces GPTScore, a metric that uses an evaluator LLM using prompts, rather than a specific functional form like the one we provide in this paper. G-EVAL, introduced in \citet{liu_g-eval_2023}, uses large language models with chain-of-thoughts (CoT) and a form-filling paradigm to determine the quality of language generation. The consideration of relevance to the question in these works is based on prompt setting, whereas our approach uses a prompt-free functional form that may be applied to the output and the query without even requiring additional context or sensitivity to the prompt. 

Perplexity is a general metric that has been used to evaluate generative text content for some time in an unsupervised manner~\citep{fomicheva_unsupervised_2020}, including code generation~\citep{xu_systematic_2022}. In addition to evaluating responses, perplexity may be used to evaluate prompts~\citep{gonen_demystifying_2022}. An alternate approach to unsupervised evaluation is to use a strong LLM as an adjudicator~\citep{bai_benchmarking_2023, liu_g-eval_2023, lin_llm-eval_2023}. 

\section{Information Metrics}
\label{sec:theory}
To a first degree, simple questions (prompts) evoke simple answers (responses), while complex questions evoke complex answers. We can measure information complexity using metrics such as entropy and its close cousin, perplexity. To recap, perplexity (PPL) is defined as the exponentiation of the entropy (in bits) of a sequence of tokens. As such, perplexity measures the number of possible binary choices (think ``20 questions game") that the combination of words implies.

Formally, assume a text sequence ${\bf X}_n$ comprising of tokens $[x_1, ..., x_n]$. Assume a performant LLM as the evaluator $\Omega$ from which sequence probabilities are obtained, i.e., this model is an oracle that returns a conditional probability $p_{\Omega}[x_t | {\bf X}_{t-1}]$. The perplexity of ${\bf X}_n$ is defined as
\begin{equation}
\label{ppl_x}
PPL({\bf X}_n) = \exp\left[-\frac{1}{n} \sum_{t=1}^n \log \left(p_{\Omega}[x_t | {\bf X}_{t-1}] \right) \right]
\end{equation}
The higher the conditional probability of the token sequence, the lower its perplexity. The evaluator LLM $\Omega$ is different from the LLM that has generated the text sequence ${\bf X}$.

Our approach in this paper uses a perplexity differential:
\begin{equation}
\label{eq:ppl_sub}
PPLqa = |PPL({\bf X}(qa)) - PPL({\bf X}(a))|, 
\end{equation} where $PPL({\bf X}(qa))$ is the perplexity of the concatenated prompt and response, and $PPL({\bf X}(a))$ is the perplexity of the response alone, as a way to evaluate the intellectual capacity of a language model. A lower value of PPLqa implies better response quality. 

This metric embodies two attributes of a good response. One, it assesses the quality of the language in the response as perplexity is designed for exactly that (language coherence and fluency). Two, since the question and answer are concatenated, it assesses the relevance and consistency of the response to the prompt, which is also an approach to detect hallucination e.g. \cite{friel_chainpoll_2023}. In order to capture relevance, we evaluate the perplexity of the response concatenated to the text of the question, $PPL({\bf X}(qa))$, from which we compute the absolute value of the difference with the perplexity of the answer, $PPL({\bf X}(a))$. If the response is unrelated to the question, then $PPL({\bf X}(qa))$ will be high. 

But even if all else is the same, perplexity may be lower for longer responses, and may be higher for more complex questions. This is usually the case for many LLMs. 
Therefore, a useful byproduct of differencing $PPL({\bf X}(qa))$ and $PPL({\bf X}(a))$ is that the differencing normalizes across responses that are of different lengths. 

The perplexity differential intuitively measures the surprise of a contextualized response given a question. A higher differential indicates a less expected response, deviating from principles like coherence and consistency. Responses adhering to these principles would be less surprising to the LLM, thus having a lower perplexity differential. 

    
    
    

The general intuition for the PPLqa metric may be summarized as an information complexity quantification of (i) a reflection of contextual understanding, (ii) an indication of coherent response, (iii) a measurement of contribution to overall perplexity, (iv) sensitivity to contextual relevance, (v) a measure of comprehensive understanding, and (vi) a model's ability to maintain topic consistency. Moreover, the metric is reproducible, practical, documented, and open, satisfying the tenets described in \citet{gu_olmes_2024}. 

\section{Procedure and Experiments}
\label{sec:experiments}
This section presents experiments to illustrate the use of the PPLqa metric and its efficacy. For objective comparison, the underlying model used in PPLqa and two other baseline methods GPTScore~\citep{fu_gptscore_2023} and G-EVAL~\citep{liu_g-eval_2023} across all experiments is Mistral 7B Instruct V0.2.\footnote{\url{https://huggingface.co/mistralai/Mistral-7B-Instruct-v0.2.}}
In particular, GPTScore applied a set of prompt templates on every question and answer pair, focusing on different aspects of evaluation such as coherence, relevance, fluency, and consistency. PPLqa by design does not require such prompts. Yet, in our experiments, we implemented PPLqa with and without such prompt templates for extensive evaluation. All experiments were run on AWS SageMaker and did not require excessive use of GPUs as no pre-training or fine-tuning was involved, only run time inference. 

GPTScore includes aspect-based prompts to sharpen the quality measurement. We applied the same prompts to PPLqa, and assessed models with and without prompts. An example of such a prompt (for the coherence aspect) is ``Answer the question based on the conversation between a human and AI. Question: Is the AI coherent and maintains a good conversation flow throughout the conversation? (a) Yes. (b) No. Conversation: human:\{question\} AI: \{answer\}.''

\subsection{Experiment I}


Our initial experiments are conducted using four different subject domains: macroeconomics, astronomy, electronics, and AI. The procedure is as follows:
\begin{enumerate}
\item Use OpenAI GPT-4 to generate 50 questions on a subject 
and also provide them in increasing order of number of bits
for all four domains. There are 200 questions in all. 
\item For each question, generate responses from different LLMs
that are popular on the Hugging Face open LLM leaderboard. 
\item For these responses, a PPLqa score is calculated using Equation~(\ref{eq:ppl_sub}) on the prompt/response pair. This score is used as a metric for a single response or for comparison across responses. 
\end{enumerate}

The PPLqa score is used to compare the responses. When the comparison is binary (i.e., responses from two LLMs are compared), the best response falls into one of two labels, and standard metrics for binary classification may be used to assess how well the metrics perform in comparison to the ground truth. Metric such as $F_1$ \citep{powers2020evaluation} score are presented for both labels. Accuracy is also presented. An important metric we use is the Matthew's correlation coefficient (MCC)~\citep{chicco2021matthews}, which  encompasses many of the simpler metrics in scope. (MCC for the binary classification case is the same as Pearson's correlation.)

\subsection{Human Evaluation of Responses}

In Experiment I, human annotators read all the responses for the subject datasets and ranked the responses from the LLMs for each question as the ground truth. 
We compare human rankings to the ones by PPLqa and other evaluating metrics.


In some cases the rankings are among binary choices, and in other cases among four choices. When the rankings are not just binary, we use rank correlations as an assessment metric. We describe this in a subsequent section titled ``Rank Correlations.''

\subsection{Results for Experiment I}

\begin{table*}
\centering
\begin{tabular}{llccccc} \toprule
Domain & Metric	 & 	PPLqa  & 	PPLqa  & 	GPTScore  & 	GPTScore 	 & 	G-EVAL	  \\ 
& & (w/prompt)	& (w/o prompt)	& (w/prompt) & (w/o prompt) \\ \midrule
Overall & $F_1$ & 0.50, 0.26 & 0.48, 0.66 & 0.46, 0.60 & 0.48, 0.67 & 0.04, 0.72  \\
& Accuracy & 0.41 & 0.59 & 0.54 & 0.6 & 0.57 \\
& MCC & -0.15 & 0.14 & 0.06 & 0.17 & 0.06 \\ 
\bottomrule
AI & $F_1$	 & 	0.65, 0.42	 & 	0.49, 0.51	 & 	0.54, 0.58	 & 	0.44, 0.55	 & 	0.07, 0.64	  \\
& Accuracy & 0.56 & 0.50 & 0.56 & 0.50 & 0.48\\
& MCC	 & 	0.096	 & 	0.010	 & 	0.135	 & 	0.023	 & 	0.132	  \\
\bottomrule									
Macroeconomics & $F_1$	 & 	0.41, 0.19	 & 	0.47, 0.77	 & 	0.43, 0.67	 & 	0.55, 0.78	 & 	0, 0.80	  \\
& Accuracy & 0.32 & 0.68 & 0.58 & 0.7 & 0.66\\
& MCC	 & 	-0.213	 & 	0.248	 & 	0.103	 & 	0.322	 & 	0.000	  \\
\bottomrule											
Astronomy & $F_1$	 & 	0.26, 0.13	 & 	0.47, 0.77	 & 	0.43, 0.59	 & 	0.56, 0.75	 & 	0.11, 0.79	  \\
& Accuracy & 0.20 & 0.66 & 0.52 & 0.68 & 0.66 \\
& MCC	 & 	-0.553	 & 	0.248	 & 	0.030	 & 	0.306	 & 	0.190	  \\
\bottomrule											
Electronics & $F_1$	 & 	0.65, 0.34	 & 	0.44, 0.55	 & 	0.42, 0.56	 & 	0.40, 0.60	 & 	0, 0.63	  \\
& Accuracy & 0.54 & 0.50 & 0.50 & 0.62 & 0.46 \\
& MCC	 & 	0.070	 & 	0.010	 & 	0.014	 & 	0.064	 & 	-0.149	  \\
\bottomrule	
\end{tabular}
\caption{\label{two_models_gt} Metrics for various evaluation methods on responses to 50 questions in four domains: AI, Macroeconomics, Astronomy, Electronics (See Experiment I). Two LLMs used to generate responses for each question are: {Titan Text G1 - Lite} and {Titan Text G1 - Express}. The comparisons here are for the various evaluation metrics against human ground truth, i.e., human labeled preference for the best answer from the two models. For metrics with two values, such as F1, the values are for both, label 0 and label 1, depending on which of the two responding LLMs is used for each label.}
\end{table*}

Table \ref{two_models_gt} shows the results for the PPLqa, GPTScore, and G-EVAL evaluation methods (with and without directed prompting) on the four domains question datasets. The domains are AI, Macroeconomics, Astronomy, and Electronics. We look at how accurately the three methods (PPLqa, GPTScore, G-EVAL) match the ranking (binary and four-way) from human evaluators as well as rankings from Claude 3. 

Results vary by domain and we evaluate based on MCC (the Matthew's Correlation Coefficient) as well as other common metrics. Overall, from Table \ref{two_models_gt}, across all four domains, PPLqa, GPTScore, and G-EVAL perform more or less the same. For the AI domain, the best method is GPTScore (w/ prompt) followed by G-EVAL, for Macroeconomics it is GPTScore (w/o prompt) followed by PPLqa (w/o prompt), for Astronomy it is GPTScore (w/o prompt) followed by PPLqa (w/o prompt), and for Electronics, it is PPLqa (w/ prompt) followed by GPTScore (w/o prompt). Depending on the domain and metric chosen, either GPTScore or PPLqa are the better evaluation methods. One of the issues is that the responses are from two similar models, and G-EVAL uses a discrete scoring method (an integer score from 1 through 5) so that in many cases it scores both responses the same and is unable to distinguish adequately between responses. This is unlikely to be an issue when the answers vary greatly in quality or style.

\begin{table*}
\centering
\begin{tabular}{llccccc} \toprule
Domain & Metric	 & 	PPLqa  & 	PPLqa  & 	GPTScore  & 	GPTScore 	 & 	G-EVAL	  \\ 
 & & (w/prompt)	& (w/o prompt)	& (w/prompt) & (w/o prompt) \\ \midrule
Overall & $F_1$ & 0.34, 0.18 & 0.54, 0.74 & 0.42, 0.64 & 0.44, 0.71 & 0.05, 0.79 \\
& Accuracy & 0.27 & 0.67 & 0.56 & 0.62 & 0.65 \\
& MCC & -0.38 & 0.28 & 0.06 & 0.15 & 0.08 \\ 
\bottomrule
AI & $F_1$	 & 	0.44, 0.19	 & 	0.57, 0.63	 & 	0.44, 0.55	 & 	0.43, 0.59	 & 	0.08, 0.69	  \\
& Accuracy & 0.34 & 0.60 & 0.50 & 0.52 & 0.54 \\
& MCC	 & 	-0.33 & 0.20 & -0.01 & 0.03 & 0.40	  \\
\bottomrule																				
Macroeconomics & $F_1$	  & 0.19. 0.13 & 0.50,  0.84 & 0.32, 0.70 & 0.44, 0.79 & 0.00, 0.88 \\
& Accuracy & 0.16 & 0.76 & 0.58 & 0.70 & 0.78 \\
& MCC	 & -0.51 & 0.35 & 0.61 & 0.26 & 0.68 \\
\bottomrule																				
Astronomy & $F_1$		& 0.34, 0.26 & 0.53, 0.76 & 0.49, 0.65 & 0.46, 0.71 & 0.11, 0.80  \\
& Accuracy & 0.30 & 0.68 & 0.58 & 0.62 & 0.68 \\
& MCC		& -0.30 & 0.29 & 0.16 & 0.17 & 0.20  \\
\bottomrule																				
Electronics & $F_1$	 & 0.38, 0.10 & 0.53, 0.71 & 0.39, 0.66 & 0.42, 0.72 & 0.00, 0.75 \\
& Accuracy & 0.28 & 0.64 & 0.56 & 0.62 & 0.60 \\
& MCC	& -0.38& 0.24 & 0.05 & 0.15 & -0.11	  \\
\bottomrule	
\end{tabular}
\caption{\label{two_models_claude} Metrics for various evaluation methods on responses to 50 questions in four domains: AI, Macroeconomics, Astronomy, Electronics (See Experiment I). Two LLMs used to generate responses for each question are: {Titan Text G1 - Lite} and {Titan Text G1 - Express}. The comparisons here are for the various evaluation metrics against labels from {Claude 3 Sonnet}. For metrics with two values, such as F1, the values are for both, label 0 and label 1, depending on which of the two responding LLMs is used for each label. }
\end{table*}

For robustness, Table \ref{two_models_claude} repeats the analysis in Table \ref{two_models_gt}, where the human labels are replaced by labels from a LLM, namely {Claude 3 Sonnet}. Overall, PPLqa is the best evaluation method. On the $F_1$ metric PPLqa (w/o prompt) does better than the others on all four domain datasets. Similar results are seen for the accuracy metric and mostly for the MCC metric as well. As before, using the scoring models without prompts usually gives better results on most metrics.

We would like to emphasize the value of not requiring a prompt in our PPLqa method, as designing prompt is a non-trivial and data dependent task. In the interest of robustness, we turn to a second experiment for further comparison of the metrics. 

\subsection{Experiment II}


The second experiment uses the MT-Bench Human Judgements dataset.\footnote{\url{https://huggingface.co/datasets/lmsys/mt_bench_human_judgments}} The questions in this dataset are provided with longer answers, to which we believe PPLqa is more  applicable. 
There are 3360 examples in the dataset. Each example contains a question, two comparable responses, and a ground truth binary label indicating which answer is better.

The GPTScore metric \citep{fu_gptscore_2023} and G-EVAL \citep{liu_g-eval_2023} metrics are used for comparison to PPLqa. A feature of the GPTScore metric (Tables 11 \& 12 in the GPTScore paper) is that it can include a prompt for evaluation as well. Hence, we used GPTScore with and without prompts, and did the same for PPLqa. G-EVAL always uses prompting. 

In MT-Bench there are two responses that are compared to determine which is better using each of the three metrics: PPLqa, GPTScore, and G-EVAL. The decisions made by each metric are compared to the ground truth binary labels introduced in the dataset. Since this is a binary classification problem, we report metrics such as precision, recall, and $F_1$ for both labels. See Table \ref{tab_mt_bench}. Results are reported for the cases with and without prompts. 

\subsection{Results for Experiment II}

\begin{table}
\centering
{\footnotesize
\label{mt_bench_results}
\begin{tabular}{lccc} \toprule
\multicolumn{4}{c}{Without prompts} \\
Metric & PPLqa & GPTScore & G-EVAL \\ \midrule
Precision & 0.510, 0.513 & 0.485, 0.497 & 0.414, 0.539\\
Recall & 0.672, 0.660 & 0.664, 0.614 & 0.910, 0.215\\
$F_1$ & 0.580, 0.577 & 0.561, 0.549 & 0.569, 0.308\\
Accuracy & 0.511 & 0.491 & 0.433 \\
MCC & 0.223 & 0.184 & 0.117\\
\bottomrule
\end{tabular}
\begin{tabular}{lccc} \toprule
\multicolumn{4}{c}{With prompts} \\
Metric & PPLqa & GPTScore & G-EVAL \\ \midrule
Precision & 0.221, 0.233 & 0.413, 0.414 & 0.414, 0.539\\
Recall & 0.285, 0.307 & 0.531, 0.547 & 0.910, 0.215\\
$F_1$ & 0.249, 0.265 & 0.465, 0.471 & 0.569, 0.308\\
Accuracy & 0.227 & 0.413 & 0.433 \\
MCC & -0.274 & 0.052 & 0.117\\
\bottomrule
\end{tabular}
}
\caption{\label{tab_mt_bench} \small Comparison of PPLqa, GPTScore, and G-EVAL on the MT-Bench Human Evaluation dataset (See Experiment II). The evaluator LLM used is {Mistral 7B Instruct V0.2}. In MT-Bench, there are two responses that are compared to determine which is better using each of the three metrics. The decisions made by each metric are compared to the ground truth binary labels
introduced in the dataset. Since this is a binary classification problem, we report metrics such as precision, recall, and F1 for both labels. For metrics with two values, such as precision, recall, F1, the values are for both, label 0 and label 1, depending on which of the two responding LLMs is used for each label.
}
\end{table}

In this experiment we compare the unsupervised evaluation methods with human evaluations on the MT-Bench Human Evaluation dataset. See Table \ref{tab_mt_bench}. The Matthew's correlation coefficient (MCC) is a good all-round metric for binary classification, and it shows that the metrics perform better without prompts. PPLqa performs best out of the three metrics. 
Specifically, PPLqa without prompts outperforms GPTScore with and without prompts as well as G-EVAL. This again highlights the advantage of PPLqa, i.e., it does not require a task dependent prompt design. In addition, when the prompts from the GPTScore paper are applied to PPLqa, PPLqa performance drops. This is because the prompts specifically designed for GPTScore are not optimized for PPLqa.

Overall, the evaluation approaches perform comparably in Experiment I and PPLqa is the best evaluation method in Experiment II. The difference in performance between the two methods across datasets is possibly difference in human annotation approaches that may be aligned differently with the various methods. 

\subsection{Experiment III}

In this experiment we re-use the data from Experiment I to explore rank correlations ($\tau$) described in Section \ref{rank_corr} below. The 200 questions in Experiment I were answered by four LLMs: (i) The  Titan Text G1 - Lite model, (ii) the Titan Text G1 - Express model, (iii) the Jurassic-2 Ultra model, and (iv) the Jurassic-2 Mid model. We use our three evaluation methods - PPLqa, GPTScore, G-EVAL - to rank all four responses to each question. Synthetic rankings are generated from the Claude 3 Sonnet model. The implementation of the above LLMs are from Amazon Bedrock\footnote{\url{https://docs.aws.amazon.com/bedrock/latest/userguide/model-ids.html}}. For each of the four evaluation domains (AI, Macroeconomics, Astronomy, Electronics) with 50 questions each, we compute Kendall's $\tau$ by ranking responses to each question. Since there are four evaluators, pairwise we have six rank correlation values for each domain and overall across domains (no prompting is applied). Next, we discuss the aggregate rank correlation measure used.

\subsection{Rank Correlations}
\label{rank_corr}

In order to compare LLMs, we examine the rankings of LLM responses to an external validation ranking, by question, averaged across all questions. This external ranking on a question may come from any validation source, such as human ranking of the responses, or ranking of the responses by a evaluator LLM. 

For example, the four subject datasets of 50 questions each were used to elicit responses from the four comparison LLMs. 
The steps in the evaluation are as follows:
\begin{enumerate}
\setlength\itemsep{0em}
\item Assume we have collected $N$ questions on a topic. For each question $q_k, k=1...N$ we get an answer $a_{kl}, \forall k$, and for each LLM $l=1...L$, where $L$ is the number of LLMs. 
\item Compute $x_{kl} = Rank(PPLqa)_{kl} \in \{1,2,...,L\}$, i.e., we rank the LLMs on each question. 
\item Collect human ranking of the LLMs responses to each question as well, denoted as $y_{kl} = Rank(Human)_{kl} \in \{1,2,...,L\}$. 
\item For each question, compute the rank correlation of PPLqa rank ($x_{kl}$) with human ranks ($y_{kl}$), denoted as $\tau_k$. 
\item Report the average of this rank correlation across all questions, i.e., 
\begin{equation}
\tau = \frac{1}{N} \sum_{k=1}^N \tau_k   \label{tau}
\end{equation}
\end{enumerate}
The higher the value of $\tau$ the greater the agreement of the PPLqa metric with human ranking. 

To compute $\tau_k$, for a question $k$, we use the standard Kendall's tau formula, where for a pair of rankings, $x_{k \cdot}$,$y_{k,\cdot}$, for $i,j \in \{1,2,...L\}$, 
\begin{equation}
\label{tau_k}
\tau_k = \frac{2}{l(l-1)} \sum_{i<j}\;sgn(x_{ki}-x_{kj}) \cdot sgn(y_{ki}-y_{kj})
\end{equation}
Here, $x_{k\cdot}$ is the first ranked vector and $y_{k\cdot}$ is the second ranked vector. There will be $l(l-1)/2$ pairwise LLM comparisons. When $sgn(x_{ki}-x_{kj})=sgn(y_{ki}-y_{kj})$, it is a ``concordant'' pair and when $sgn(x_{ki}-x_{kj}) \neq sgn(y_{ki}-y_{kj})$, then it is a ``discordant'' pair. Rank correlation is the net proportion of concordant(positive) and discordant (negative) pairs. 

From the formula, it is easy to see that if the sort order of $x_{k\cdot}$ agrees with that of $y_{k\cdot}$, then the sign of the in-sum term is $+1$, and it is $-1$ for reversed rankings. It follows that if the ranks for $x$ and $y$ columns are identical, then $\tau_k=1$. As with correlations, $-1 \leq \tau_k \leq +1$. The same is true for the aggregate metric $\tau$.

\subsection{Using a large LLM as evaluator}
In Experiment III, we prepare ground truth from LLM evaluations. For each question, we submit all $L$ responses from LLMs to an ``evaluator'' LLM, and we use Claude 3 from Anthropic as the evaluator. It has a large context window ($>100$K) and can accommodate responses from multiple LLMs concatenated together. 

Prompt engineering experimentation resulted in the following prompt that asks Claude 3 to rank four responses:
\begin{verbatim}
promptq = """Human: Rank the following 4 
Answers based on the Question. \
The response should contain the answers 
in ranked order from best to worst. \
    Question: {{question}} \
    Answer 1: 'ANSWER_1' \
    Answer 2: 'ANSWER_2' \
    Answer 3: 'ANSWER_3' \
    Answer 4: 'ANSWER_4' \
    Assistant:        """
\end{verbatim}
where {\tt ANSWER\_*} contains the four respective responses that are to be ranked. This prompt delivers the ranking as well as the reasoning. {\tt regex} is used to extract the ranking only from among the detailed text response.





\subsection{Comparison to Rank Correlations on Hugging Face Leaderboard}

The $\tau$ metric assesses how consistently the  PPLqa metric ranks LLMs versus external rankings by humans or LLMs. We may adapt the same approach to assess how consistent the rankings are on a leaderboard across different tests/datasets. This gives an idea of how large or small the $\tau$ value may be empirically across evaluation methods.

Take the Hugging Face leaderboard as an example (as on 27-September-2023). 
The data comprises all LLMs that had average scores greater than 70. This yielded 73 LLMs. The test scores on four datasets (ARC, HellaSwag, MMLU, TruthfulQA) are shown as well as the average score, which is used for ranking. Since there are four tests, rank correlations may be computed pairwise for all pairs of LLMs, six pairs in total. These six rank correlations are averaged to give a single measure of how consistently different tests rank the various LLMs. 

Kendall's tau (averaged across six pairs of tests) for all 73 LLMs is 0.50. If $\tau$ is computed using only the top 10 LLMs of the leaderboard, then $\tau=0.43$. For just top 4 LLMs of the leaderboard, $\tau=0.33$. Therefore, the average rank correlation depends on how many ranks we are evaluating and increases with the number of ranking rows on the Hugging Face leaderboard. When the number of ranking rows is four, the same as $L$, the number of LLMs in our experiment, the $\tau$ value of 0.33 is coincidentally similar to that obtained when comparing the ranking of PPLqa with human model rankings (macroeconomics) and slightly higher than that for the other subject datasets. However, we may conclude from the $\tau$ value that the Hugging Face leaderboard shows consistency across the various tests applied to open LLMs. 

The level of consistency may also be compared to a baseline generated by Monte Carlo simulation as follows: Generate two ranked lists of four items each and compute Kendall's tau. Repeat this a large number of times, in our case we ran 100K repetitions. Assuming conditioning on some positive rank correlation of humans and machines, we discard all values less than 0 and compute the mean of the remaining tau values. This is 0.31, which is a baseline for what we get with four item lists conditional on the fact that there is some ranking ability. This number also compares closely to what we get in our experiments.


\subsection{Results for Experiment III}

\begin{table}
\centering
{\small
\noindent\begin{tabular}{p{0.4in}cccc}
\toprule
Domain & PPLqa & GPTScore & G-EVAL & {GPTScore} \\
& w/Claude & w/Claude & w/Claude & {w/PPLqa} \\
\midrule
Overall & 0.137 & 0.197 & -0.074 & 0.199 \\
 & 0.01 & 0.00 & 0.21 & 0.61 \\ \midrule
AI & 0.049 & 0.102 & -0.062 & 0.192 \\
 & 0.64 & 0.32 & 0.60 & 0.06 \\ \midrule
Macro- & 0.234 & 0.400 & -- & 0.106 \\
 economics & 0.02 & 0.00 & -- & 0.30 \\ \midrule
Astronomy & 0.126 & 0.266 & -0.021 & 0.176 \\
 & 0.22 & 0.01 & 0.86 & 0.09 \\ \midrule
Electronics & 0.026 & -0.003 & -0.149 & 0.319 \\
 & 0.80 & 0.97 & 0.21 & 0.00 \\ 
\bottomrule
\end{tabular}
}
\caption{\label{rank_corr_results} \small Kendall's $\tau$ for pairwise correlation in model rankings. Correlation is computed across rankings by: PPLqa, GPTScore, and G-EVAL, as well as a synthetic proxy for ground truth, {Claude 3 Sonnet}. These rankings are across the four evaluation domains - AI, Macroeconomics, Astronomy, Electronics - with 50 questions each (See Experiment III). The 200 questions in Experiment I were answered by four LLMs: (i) {Titan Text G1 - Lite}, (ii) {Titan Text G1 - Express}, (iii) {Jurassic-2 Ultra}, and (iv) {Jurassic-2 Mid}. No prompting is applied. Each domain below had two rows: the first row reports the Kendall's tau and the second row reports the corresponding $p$-value.
}
\end{table}

The same data is used as in Experiment I, but with four LLMs responses instead of binary ranking only. Table \ref{rank_corr_results} presents the average rank correlation (Kendall's tau) for each question and four responses across different evaluator models with the rankings from Claude 3. The rank correlation of GPTScore with PPLqa is also shown. GPTScore evidences the highest rank correlations except for one domain. G-EVAL does not have high rank correlation in all domains. The differences in responses across LLMs is not really stark, as a reading of many sets of responses shows, and this also explains why rank correlations are on the lower side. Finally, the overall average rank correlation between PPLqa and GPTScore (without prompts) is around 20\% overall, in the ballpark of correlations with the Claude rankings. This is caveated by the fact that these correlations are with a model's ranking and not human rankings as shown in Table \ref{tab_mt_bench}.

\section{Limitations of the Approach}
\label{sec:limits}

PPLqa is a simple prompt-free metric with a quick procedure that is able to achieve good correlation with human rankings. We propose it as a simple measure of quality to compare different LLMs in the absence of ground truth and expert evaluators. 
The scope of our work is deliberately narrow, aimed at rankings of LLMs in order to help a human select which one would be the best candidate for use on a certain task or for further tuning. The only input required is a series of questions with answers from multiple LLMs, and no ground truth.  Further, we focus on responses that are long form text and not responses to binary/multiple choice questions that generally fall within the domain of classifier models. The information-theoretic interpretation of such models is explained in~\citet{friedland_information-driven_2024}. We expect that our metric does not reflect every nuance that one might want to capture in a given, specific task. It will also be less useful when ground truth is available and traditional approaches are easy to apply. On the plus side, no language specifics are used, so the metric can work with any trained language. The range of datasets to which the metric is applied is limited. Other than GPTScore, other metrics are less comparable and have different foundations or use advanced prompting. No quality measure is perfect, neither is ours, and so accuracy will always continue to be a point of discussion for any method.

\section{Conclusions}
\label{sec:concl}
Unlike other metrics for LLM evaluation that need costly ground truth for evaluation, we propose a metric based on information theory that is unsupervised and requires no ground truth. We choose four example domains for illustrating our evaluation metric through various experiments. We also apply the metric to the MT-Bench human evaluation dataset for comparison with other unsupervised metrics such as GPTScore and G-EVAL. 
    
First, we argue that our metric captures both the language quality of the response as well as its relevance to the query. Second, to illustrate application of the metric we undertake a comparison of  LLM responses. Examples are provided so that human evaluation is seen to be consistent with our PPLqa metric. Third, PPLqa also correlates positively to ranking by an LLM (Claude 3).  
We propose the method discussed in here as a general measure of quality to compare different LLMs in the absence of ground truth and expert evaluators.

We note that our metric is a ``white-box'' metric, in that it is fully transparent in the computations made and one may investigate each neuron's activation in the softmax layer from which the perplexity metric is computed. This is different from black-box methods that rely on querying the opinion of an LLM for ranking answers, such as the G-EVAL one used for comparison in this paper. 
 
Further work can explore a theoretical analysis of the metric, possibly using a theoretic model of an LLM. It may also be useful to develop a series of domain specific test banks to support a PPLqa leaderboard. 



\bibliography{DeepLearning_LLM_ContinualLearning}


\clearpage
\onecolumn

\appendix

\begin{center}
    {\Large {\bf Appendix: Supplementary Material}}
\end{center}

\begin{figure*}
\centering
\caption{\small Sample questions generated by a LLM for four topics: AI, astronomy, electronics, and macroeconomics.}
\label{fig:gpt4q}
\includegraphics[scale=0.35]{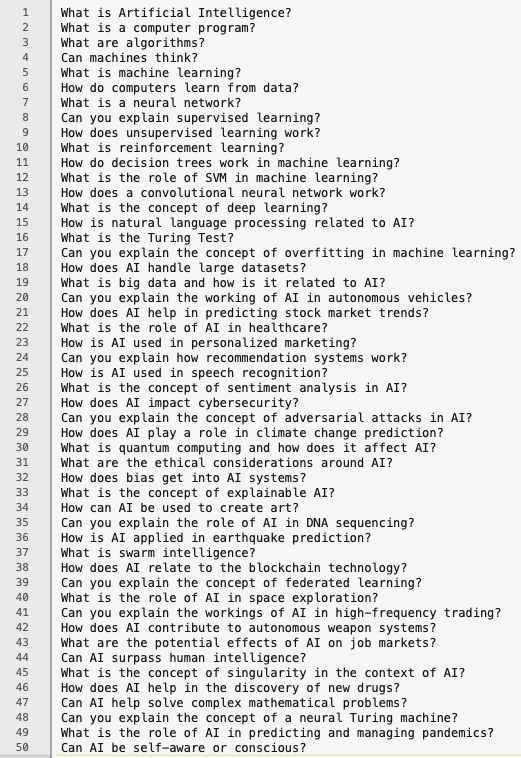}
\includegraphics[scale=0.35]{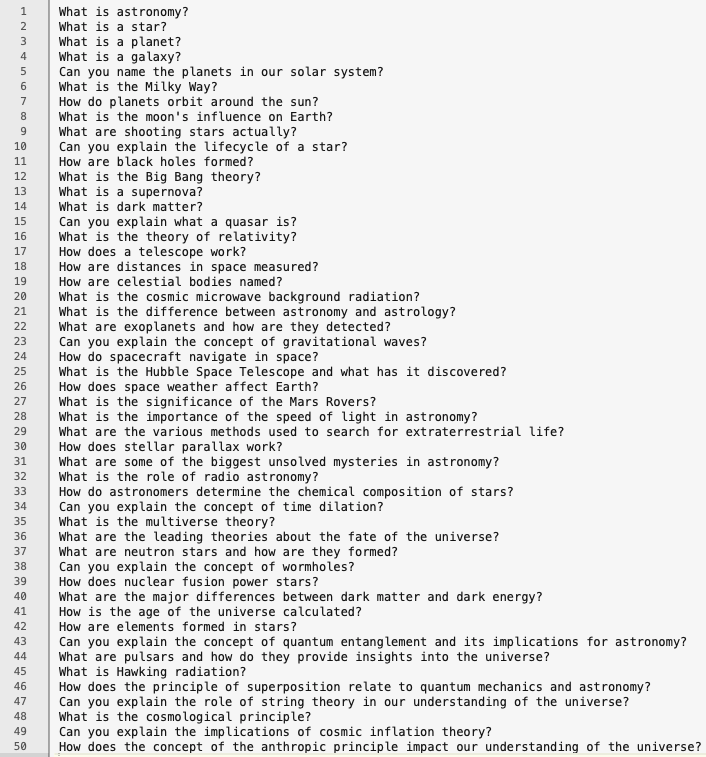}
\includegraphics[scale=0.34]{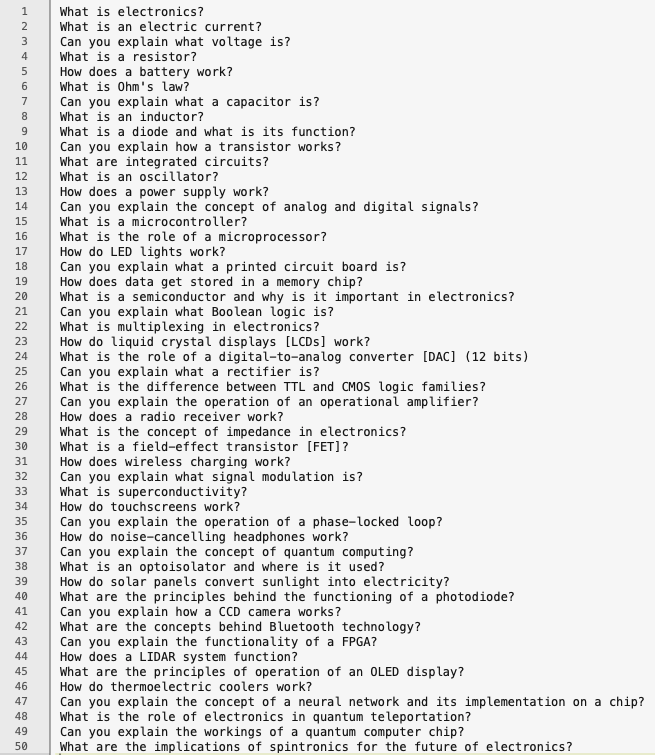}
\includegraphics[scale=0.34]{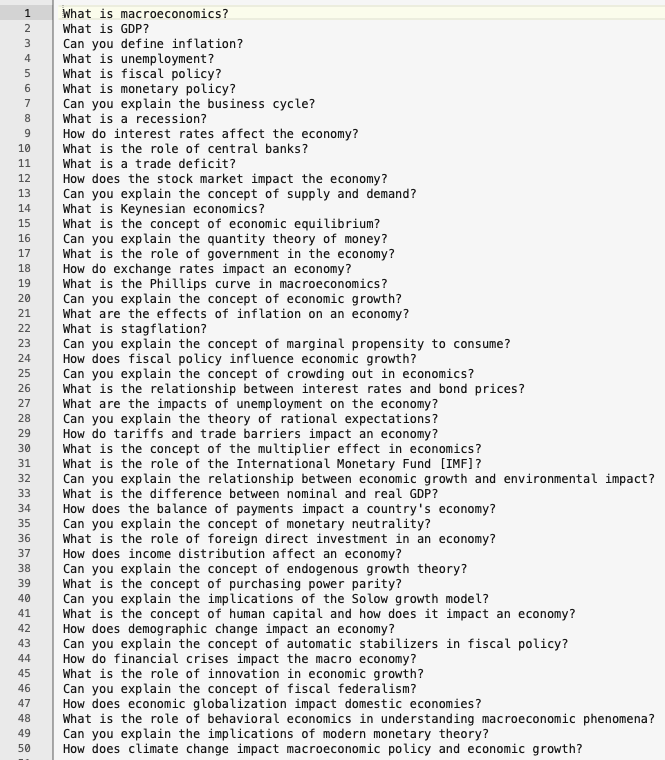}
\end{figure*}



\begin{table*}[h]
\centering
\caption{\label{ppl_len_corr} \small Correlation of perplexity $PPL({\bf X}(qa))$ with the number of words in the response. Verifying that perplexity tends to be smaller when the length of text is greater.
}
\begin{tabular}{lcccc} \toprule
Model & Macroeconomics & Astronomy & AI & Electronics\\ \midrule
Samantha-1.11-70b & -0.172 & -0.306 & -0.275 & -0.258\\
ORCA\_LLaMA\_70B\_QLoRA &  -0.171 & -0.197 & -0.395 & -0.185\\
Uni-TianYan & -0.283 & -0.360 & -0.019 & -0.238\\ 
samantha-1.1-llama-33b & -0.335 & -0.323 & 0.232 & -0.191\\
\bottomrule
\end{tabular}
\end{table*}

\end{document}